\documentclass[runningheads]{llncs}
\usepackage{graphicx}
\usepackage{amsmath}
\usepackage{booktabs}
\usepackage{amssymb}
\usepackage{comment}
\usepackage{wrapfig}
\usepackage{todonotes}
\usepackage{multirow}
\usepackage{bm}
\usepackage{hyperref}
\begin{document}
\title{Segmenting two-dimensional structures with strided tensor networks}
%\thanks{Supported by organization x.
\titlerunning{Strided Tensor Networks}
\author{Raghavendra Selvan\inst{1,2}
% \author{First Author \inst{1,2}
%\orcidID{} 
\and
Erik B Dam\inst{1}
% Second Author\inst{1} 
\and
Jens Petersen\inst{1,3}
% Third Author\inst{1,3}
}
\authorrunning{R. Selvan et al.}
% \authorrunning{First Author et al.}
% \institute{Institute 1 \and
% Institute 2 \and 
% Institute 3\\
% \email{mail@email}}
\institute{Department of Computer Science, University of Copenhagen 
\and Department of Neuroscience, University of Copenhagen
\and Department of Oncology, Rigshospitalet, Denmark\\
\email{raghav@di.ku.dk}}
\def \Em{{\mathbb{E}}}
\def \Rm{{\mathbb{R}}}

\def \Im{{\mathbb{I}}}
\def \abf{{\mathbf a}}
\def \Abf{{\mathbf A}}
\def \bbf{{\mathbf b}}
\def \Bbf{{\mathbf B}}
\def \cbf{{\mathbf c}}
\def \Cbf{{\mathbf C}}
\def \dbf{{\mathbf d}}
\def \Dbf{{\mathbf D}}
\def \ebf{{\mathbf e}}
\def \Ebf{{\mathbf E}}
\def \fbf{{\mathbf f}}
\def \Fbf{{\mathbf F}}
\def \gbf{{\mathbf g}}
\def \Gbf{{\mathbf G}}
\def \hbf{{\mathbf h}}
\def \Hbf{{\mathbf H}}
\def \ibf{{\mathbf i}}
\def \Ibf{{\mathbf I}}
\def \jbf{{\mathbf j}}
\def \Jbf{{\mathbf J}}
\def \kbf{{\mathbf k}}
\def \Kbf{{\mathbf K}}
\def \lbf{{\mathbf l}}
\def \Lbf{{\mathbf L}}
\def \mbf{{\mathbf m}}
\def \Mbf{{\mathbf M}}
\def \nbf{{\mathbf n}}
\def \Nbf{{\mathbf N}}
\def \obf{{\mathbf o}}
\def \Obf{{\mathbf O}}
\def \pbf{{\mathbf p}}
\def \Pbf{{\mathbf P}}
\def \qbf{{\mathbf q}}
\def \Qbf{{\mathbf Q}}
\def \rbf{{\mathbf r}}
\def \Rbf{{\mathbf R}}
\def \sbf{{\mathbf s}}
\def \Sbf{{\mathbf S}}
\def \tbf{{\mathbf t}}
\def \Tbf{{\mathbf T}}
\def \ubf{{\mathbf u}}
\def \Ubf{{\mathbf U}}
\def \vbf{{\mathbf v}}
\def \Vbf{{\mathbf V}}
\def \wbf{{\mathbf w}}
\def \Wbf{{\mathbf W}}
\def \xbf{{\mathbf x}}
\def \Xbf{{\mathbf X}}
\def \ybf{{\mathbf y}}
\def \Ybf{{\mathbf Y}}
\def \zbf{{\mathbf z}}
\def \Zbf{{\mathbf Z}}
\def \Hbf{{\mathbf H}}
\def \0bf{{\mathbf 0}}

\def \Emean{\mathbb{E}}
\def \Rm{\mathbb{R}}

\def \Sibf{{\mathbf \Sigma}}
\def \xbbf{\mathbf{\bar{x}}}
\def \etr{\mbox{etr}}
\def \tr{\mbox{tr}}
\def \Tr{\mbox{Tr}}
\def \Cov{\mbox{Cov}}
\def \cost{\mbox{cost}}
\def \diag{\mbox{diag}}
\def \Lambf{{\mathbf{\Lambda}}}
\def \Gambf{{\mathbf{\Gamma}}}
\def \Sigbf{{\mathbf \Sigma}}
\newcommand{\rhobf}{\ensuremath{\boldsymbol{\rho}}}
\newcommand{\lambf}{\ensuremath{\boldsymbol{\lambda}}}
\newcommand{\nubf}{\ensuremath{\boldsymbol{\nu}}}
\def \alphabf{{\boldsymbol{\alpha}}}
\def \sigmabf{{\boldsymbol{\sigma}}}
\def \mubf{{\boldsymbol{\mu}}}

\def \Ncal{{\mathcal N}}
\def \Pcal{{\mathcal P}}
\def \Fcal{{\mathcal F}}
\def \Ecal{{\mathcal E}}
\def \Scal{{\mathcal S}}
\def \Qcal{{\mathcal Q}}
\def \Bcal{{\mathcal B}}
\def \Lbcal{{\mathcal Lb}}
\def \Gcal{{\mathcal G}}
\def \Lcal{{\mathcal L}}
\def \KLD{\text{KLD}}
\def \KL{\text{KL}}

%\\
\maketitle              % typeset the header of the contribution
\begin{abstract}
\vspace{-0.5cm}
Tensor networks provide an efficient approximation of operations involving high dimensional tensors and have been extensively used in modelling quantum many-body systems. More recently, supervised learning has been attempted with tensor networks, primarily focused on tasks such as image classification. In this work, we propose a novel formulation of tensor networks for supervised image segmentation which allows them to operate on high resolution medical images. We use the matrix product state (MPS) tensor network on non-overlapping patches of a given input image to predict the segmentation mask by learning a pixel-wise {\em linear} classification rule in a high dimensional space. The proposed model is end-to-end trainable using backpropagation. It is implemented as a {\em strided tensor network} to reduce the parameter complexity.  The performance of the proposed method is evaluated on two public medical imaging datasets and compared to relevant baselines. The evaluation shows that the strided tensor network yields competitive performance compared to CNN-based models while using lower resources such as GPU memory. Additionally, based on the experiments we discuss the feasibility of using fully linear models for segmentation tasks.
%Tensor networks provide an efficient approximation of operations involving high dimensional tensors and have been extensively used in modelling quantum many-body systems. More recently, supervised learning has been attempted with tensor networks, primarily focused on tasks such as image classification. In this work, we propose a novel formulation to apply tensor networks for supervised image segmentation allowing them to operate on high resolution images. We use the matrix product state (MPS) tensor network on non-overlapping patches of a given input image to predict the segmentation mask by learning a pixel-wise {\em linear} classification rule in a high dimensional space. The proposed model is end-to-end trainable using backpropagation. It is implemented as a {\em strided tensor network} to reduce the parameter complexity.  The performance of the strided tensor network is evaluated on two public medical imaging datasets and compared to relevant baseline methods. The evaluation shows that the proposed yields competitive performance compared with CNN-based models while using lower resources such as GPU memory. With the experiments we also provide insights into the feasibility of using linear models like tensor networks for segmentation purposes.
\footnote{Source code: \url{https://github.com/raghavian/strided-tenet}}

\vspace{-0.25cm}

\vspace{-0.5cm}
\end{abstract}

% {\bf For submission:}

% \begin{enumerate}
% \item    Submit an abstract of your paper.
% \item Summarize the contributions of your paper.
% \item Describe in 3 sentences or less what are the medical objectives of your paper.
% \item Describe in 3 sentences or less what you think is most methodologically interesting about your paper.
% \item Describe in 3 sentences or less what discussions your paper could create at IPMI.
% \end{enumerate}

% {\bf High level overview of the paper}
% \begin{enumerate}
%     \item Linear classification of pixels in high dimensional spaces
%     \item Similar to patch based methods
%     \item Multiple datasets: Lung CXR, Retina, LIDC (not great results)
%     \item Influence of augmentation to study if MPS learns location specific strucutres (It does not!)
%     \item Conceptual comparison with SVMs, MLPs, CNNs, Random Ferns?
%     \item More sensitive to feature dimension than bond dimension!
%     \item Influence of feature dimensions and intuition is important. In MLTN also larger squeeze neighbourhoods yielded better performance. Other than increasing the space dimensionality, what else does this do? 
%     \item nJet: Interestingly just the sinusoidal map is as good as jet features (sometimes even better for Retina dataset)
%     \item What about non-MPS models? PEPS discussion
% \end{enumerate}

\vspace{-0.25cm}
\section{Introduction}
\vspace{-0.25cm}

%Image segmentation has been studied for decades. 
Large strides made in the quality of computer vision in the last decade can be attributed to deep learning based methods~\cite{lecun2015deep}; in particular, to the auxiliary developments (powerful hardware, better optimisers, tricks such as dropout, skip connections etc.) that have made convolutional neural networks (CNNs) more effective.
%~\cite{lecun1989backpropagation}. 
This has also influenced biomedical image segmentation with models such as the U-net~\cite{ronneberger2015u} which have become widely popular\footnote{First author of~\cite{ronneberger2015u} noted their U-net work was cited more than once every hour in 2020. \url{https://bit.ly/unet2020}}.

% However, the dependency of CNN-based models on high quality labeled data and specialized hardware could make them less appealing in certain situations. For instance, when dealing with biomedical images where labels are expensive; or in developing countries where access to expensive hardware might be scarce~\cite{ahmed2020democratization}. The carbon footprint associated with training large deep learning models is also growing to become a concern~\cite{anthony2020carbontracker}. Investigating fundamental ideas that can alleviate some or all of these concerns could be valuable going forward. It is in this spirit that we here explore the possibility of using tensor networks for image segmentation. 

Tensor networks are factorisations of high dimensional tensors, and have been widely used to study quantum many-body systems~\cite{orus2014practical}. Conceptually, they can be interpreted as linear models operating in high dimensional spaces, in contrast to neural networks which are highly non-linear models operating in lower dimensional spaces. Tensor networks have been used as feature extractors~\cite{bengua2015matrix}, predictors operating in very high dimensional spaces~\cite{novikov2018exponential} and to compress neural networks~\cite{novikov2015tensorizing}. More recently, they are also being studied in the context of supervised learning with growing success~\cite{stoudenmire2016supervised,efthymiou2019tensornetwork,raghav2020tensor}. They have been primarily used for image classification~\cite{stoudenmire2016supervised,efthymiou2019tensornetwork} and most recently to classify medical images~\cite{raghav2020tensor}. Tensor networks have not been studied  for image segmentation to the best of the authors' knowledge.

In this work, we propose the strided tensor network: a tensor network based image segmentation method. Tensor network operations are performed on image patches to learn a hyper-plane that classifies pixels into foreground and background classes in a high dimensional space. This is similar to classical pixel classification methods operating in some expressive feature space~\cite{soares2006retinal,vermeer2011automated}. The key difference with tensor networks is that they do not require designed features that encode domain specific knowledge, and still are able to learn {\em linear} models that are competitive with state-of-the-art CNN-based models, as has been shown for tasks such as image classification~\cite{efthymiou2019tensornetwork,raghav2020tensor}. Further, the proposed model can be trained in an end-to-end manner in a supervised learning set-up by backpropagating a relevant loss function. We experiment on two biomedical imaging datasets: to segment nuclei from microscopy  images of multi-organ tissues, and to segment lungs from chest X-rays (CXR). We compare the strided tensor network with relevant baselines, including deep learning models, and demonstrate that the tensor network based model can yield similar performance compared to CNN-based models with fewer resources.

%Strategies such as data augmentation and self-supervision are being explored to alleviate these issues within the CNN framework are being studied~\cite{perez2017effectiveness}. 

% \cite{vermeer2011automated,soares2006retinal,shi2016real,larsen2012jet,stoudenmire2016supervised,efthymiou2019tensornetwork,raghav2020tensor,orus2019tensor}

\vspace{-0.25cm}
\section{Methods}
\vspace{-0.25cm}
\subsection{Overview}
In this work, we propose a tensor network based model to perform image segmentation. This is performed by approximating the segmentation decision as a {\em linear} model in an exponentially high dimensional space. That is, we are interested in deriving a hyper-plane in a high dimensional space such that it is able to classify pixels into foreground and background classes across all images in the dataset.

\begin{figure}[t]
    \centering
    \includegraphics[width=0.8\textwidth]{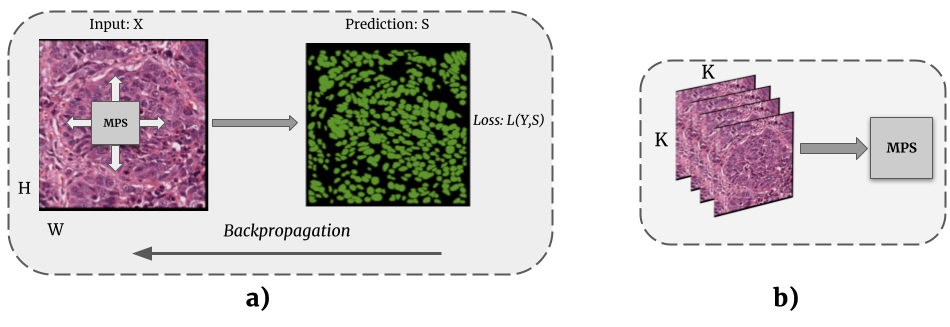}
    \vspace{-0.25cm}
    \caption{{\bf a)} High level overview of the proposed strided tensor network model. Matrix product state (MPS) operations are performed on non-overlapping regions of the input image, $X$, of size H$\times$W resulting in the predicted segmentation, $S$. Loss computed between prediction, $S$, and the ground truth, $Y$, are backpropagated during training. Arrows on MPS block are used to indicate that the same MPS block is applied across the image. {\bf b)} In practice, the strided tensor network operations can be accelerated using batch processing by creating non-overlapping patches (size K$\times$K) on the fly as input to MPS, and tiling these batched predictions to reconstruct the full image segmentation.} 
    %We use stride kernels of $K=8$ \& $32$ for two different datasets (Section~\ref{sec:data_exp}).}
    \label{fig:model}
    \vspace{-0.25cm}
\end{figure}

We consider non-overlapping image patches, flatten them into 1D vectors and apply simple sinusoidal feature transformations resulting in {\em local} feature maps. By taking tensor product of these local feature maps, we obtain {\em global} feature maps which in effect lift the input image patch into an exponentially high dimensional space. Weights for the linear model that operate on the global feature maps, resulting in segmentation predictions, are approximated using the matrix product state (MPS) tensor network\footnote{Matrix product state tensor networks are also known as Tensor Trains in literature.}~\cite{perez2006matrix,oseledets2011tensor}. The same trainable MPS is used on non-overlapping image patches from the entire image resulting in our strided tensor network model for image segmentation. Predicted segmentations are compared with ground truth labels in the training set to obtain a suitable loss which is backpropagated to optimise the weights of our model. A high level overview of the proposed model is illustrated in Figure~\ref{fig:model}.

In the remainder of this section, we present a brief introduction to tensor notation, describe the choice of local feature maps, going from local to global feature maps and details on approximating the segmentation rule with MPS, in order to fully describe the proposed strided tensor network for image segmentation.

% \vspace{-0.25cm}
\subsection{Tensor notation}

\begin{figure}[t]
    \centering
    \includegraphics[width=0.9\textwidth]{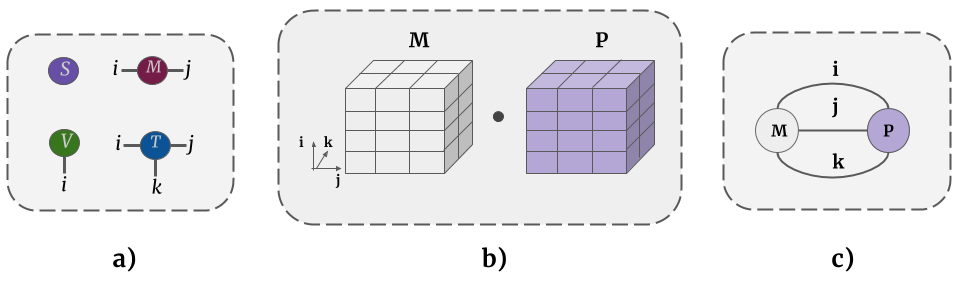}
    \vspace{-0.25cm}
    \caption{ {\bf a)} Graphical tensor notation depicting an order-0 tensor (scalar $S$), order-1 tensor (vector $V^i$), order-2 tensor (matrix $M^{ij}$) and an order-3 tensor $T^{ijk}$. Tensor indices -- seen as the dangling edges -- are written as superscripts by convention. {\bf b)} Illustration of $<M,P>$: the dot product of two order-3 tensors of dimensions $i=4,j=3,k=2$. {\bf c)} Tensor notation depicting the same dot product between the tensors $M_{ijk}$ and $P_{ijk}$. Indices that are summed over (tensor contraction) are written as subscripts.  Tensor notations can capture operations involving higher order tensors succinctly.}
    \label{fig:tensor}
    \vspace{-0.25cm}
\end{figure}

Tensor notations are concise graphical representations of high dimensional tensors introduced in~\cite{penrose1971applications}. A grammar of tensor notations has evolved through the years enabling representation of complex tensor algebra. This not only makes working with high dimensional tensors easier but also provides insight into how they can be efficiently manipulated. Figure~\ref{fig:tensor} shows the basics of tensor notations and one important operation -- tensor contraction (in sub-figures b \& c). We build upon the ideas of tensor contractions to understand tensor networks such as the MPS, which is used extensively in this work. For more detailed introduction to tensor notations we refer to~\cite{bridgeman2017hand}.
%\footnote{\url{https://tensornetwork.org/} also has some well-curated introductory material.}

\vspace{-0.25cm}
\subsection{Image segmentation using linear models}

In order to motivate the need for using patches, we first describe the model at the full image level. 

Consider a 2 dimensional image, $X \in \Rm^{H\times W \times C}$ with $N=H\times W$ pixels and $C$ channels. The task of obtaining an $M$--class segmentation, $Y \in \{0,1\}^{H\times W \times M}$ is to learn the decision rule of the form $f (\cdot\;;\;\Theta): X \mapsto Y$, which is parameterised by $\Theta$. These decision rules, $f(\cdot\;;\;\Theta)$, could be non-linear transfer functions such as neural networks. In this work, building on the success of tensor networks used for supervised learning~\cite{stoudenmire2016supervised,efthymiou2019tensornetwork,raghav2020tensor}, we explore the possibility of learning $f(\cdot\;;\;\Theta)$ that are linear. For simplicity, we assume two class segmentation of single channel images, implying $M=1$, $C=1$. However, extending this work to multi-class segmentation of inputs with multiple channels is straightforward.

Before applying the linear decision rule, the input data is first lifted to an exponentially high dimensional space. This is based on the insight that non-linearly separable data in low dimensions could possibly become linearly separable when lifted to a sufficiently high dimensional space~\cite{cortes1995support}. The lift in this work is accomplished in two steps.

First, the image is flattened into a 1-dimensional vector $\xbf \in \Rm^N$. Simple transformations are applied to each pixel to increase the number of features per pixel. These increased features are termed local feature maps.
%The local feature map for a pixel $x_j$ is given as $\psi^{i_j}(x_j): \Rm \mapsto \Rm^d$. 
% In quantum physics applications, simple sinusoidal transformations that are related to wavefunction superpositions are used as the local feature maps~\cite{stoudenmire2016supervised}. 
Pixel intensity-based local feature maps have been explored in recent machine learning applications of tensor networks~\cite{efthymiou2019tensornetwork,reyes2020multi,raghav2020tensor}.  We use a general sinusoidal local feature map from~\cite{stoudenmire2016supervised}, which increases local features of a pixel from $1$ to $d$: % i.e, $\psi^{i_j}(x_j): [0,1] \mapsto [0,1]^d$ given as
\begin{equation}
\psi^{i_j}(x_j) =  \sqrt{{d-1 \choose i_j-1}} \left(\cos(\frac{\pi}{2}x_j)\right)^{(d-i_j)}\left(\sin(\frac{\pi}{2}x_j)\right)^{(i_j-1)}
{ \forall \; i_j=1\dots d}. \label{eq:local}
\end{equation}
The intensity values of individual pixels, $x_j$, are assumed to be normalised to be in $[0,1]$. Further, the local feature maps are constrained to have unit norm so that the global feature map in the next step also has unit norm.

% which is then flattened into a 1-dimensional vector $\xbf \in \Rm^N$. This flattened input image is lifted into a high dimensional space in two steps: first, a pixel-level {\em local} feature map is applied to increase the feature dimension. For any pixel, $x_j$, it is given by $\psi^{i_j}(x_j): \Rm \mapsto \Rm^d$. Commonly used feature maps in literature include sinusoidal or intensity transformations~\cite{stoudenmire2016supervised,efthymiou2019tensornetwork,reyes2020multi}. 
In the second step, a global feature map is obtained by taking the tensor product\footnote{Tensor product is the generalisation of matrix outer product to higher order tensors.} of the local feature maps. This operation takes $N$ %$d$--dimensional local map tensors of
order-1 tensors and outputs an
%$d$--dimensional
order-$N$ tensor, $\Phi^{i_1\dots i_N}(\xbf) \in [0,1]^{d^N}$ given by
\begin{equation}
    \Phi^{i_1\dots i_N}(\xbf) = \psi^{i_1}(x_1) \otimes \psi^{i_2}(x_2) \otimes \dots \otimes \psi^{i_N}(x_N).
    \label{eq:joint}
\end{equation}
Note that after this operation each image can be treated as a vector in the $d^N$ dimensional Hilbert space~\cite{orus2014practical,stoudenmire2016supervised}.

Given the $d^N$ global feature map in Equation~\eqref{eq:joint}, a linear decision function $f(\cdot;\Theta)$ can be estimated by simply taking the tensor dot product of the global feature map with an order-(N+1) weight tensor, $\Theta^{m}_{i_1\dots i_N}$:
\begin{equation}
    f^m(\xbf\;;\;\Theta) =  \Theta^{m}_{i_1\dots i_N}  \cdot  \Phi_{i_1\dots i_N}(\xbf).
    \label{eq:linModel}
\end{equation}
The additional superscript index $m$ on the weight tensor and the prediction is the output index of dimension $N$. That is, the resulting order-1 tensor from Equation~\eqref{eq:linModel} has $N$ entries corresponding to the pixel level segmentations. Equation~\eqref{eq:linModel} is depicted in tensor notation in Figure~\ref{fig:mps}-a.

% Sequence of equations corresponding to local and global feature maps, and the linear decision model are given below.
% \begin{align}
% \textbf{Local map: }&& \psi^{i_j}(x_j) &&=& \quad \sqrt{\frac{(l-1)}{(s-1)}} \left(\cos(\frac{\pi}{2}x_j)\right)^{(l-s)}\left(\sin(\frac{\pi}{2}x_j)\right)^{(s-1)} \nonumber \\  &&&&&\qquad \qquad \qquad \qquad\qquad \qquad 
% { \forall\; l=1\dots d} \label{eq:local} \\
% \textbf{Global map: }&&    \Phi^{i_1\dots i_N}(\xbf) &&=&\quad \psi^{i_1}(x_1) \otimes \psi^{i_2}(x_2) \otimes \dots \otimes \psi^{i_N}(x_N) 
%     \label{eq:joint} \\
% \textbf{Linear model: }&& f_\Theta(\xbf) &&=&\quad  \Theta^{m}_{i_1\dots i_N}  \cdot  \Phi_{i_1\dots i_N}(\xbf)
%     \label{eq:linModel}
% \end{align}
% \vspace{-0.5cm}
\subsection{Strided tensor networks}
\vspace{-0.15cm}
% where,
% \begin{equation}
% \psi^{i_j}(x_j;k) = \sqrt{\frac{(l-1)}{(s-1)}} (\cos(\frac{\pi}{2}x_j))^{(l-s)}(\sin(\frac{\pi}{2}x_j))^{(s-1)} \qquad  \forall\; l=1\dots k
% %\{ X \in \Rm^{H \times W \times d},d=1 \}  \longrightarrow \{\xbf \in \Rm^{ (N/k^2) \times d}, d= k^2\}.
% \label{eq:squeeze}
% \end{equation}
% $l$ runs from $1$ to $k$. Given the high dimensional joint feature map\footnote{The tensor indices ${i_1\dots i_N}$ are dropped for ease of notation.}, $\Phi(\xbf)$, the linear  decision boundary is given by the following tensor inner product:
% \begin{equation}
%     f^m(\xbf) = \left ( \Theta^{m}_{i_1\dots i_N}(\xbf) \right) \cdot \left( \Phi_{i_1\dots i_N}(\xbf) \right),
%     \label{eq:linModel}
% \end{equation}
% where $\Theta$ is an order-($N+1$) weight tensor, with output dimension, $m$, corresponding to the number of output classes\footnote{We show the indices that are being summed over as subscripts following tensor contraction notation.}. 

%
While the dot product in Equation~\eqref{eq:linModel} looks easy enough conceptually, on closer inspection its intractability comes to light. The approach used to overcome this intractability leads us to the proposed strided tensor network model.
% \\ \\
%%
\begin{enumerate}
    \item {\bf Intractability of the dot product:} The sheer scale of the number of parameters in the weight tensor $\Theta$ in Equation~\eqref{eq:linModel} can be mind boggling. For instance, the weight tensor required to operate on an tiny input image of size 32$\times$32 with local feature map $d=2$ is $N\cdot d^N=1024 \cdot 2^{1024}\approx 10^{79}$ which is close to the number of atoms in the observable universe\footnote{\url{https://en.wikipedia.org/wiki/Observable_universe}} (estimated to be about $10^{80}$). 
    \item {\bf Loss of spatial correlation:} The assumption in Equations~\eqref{eq:local},~\eqref{eq:joint} and~\eqref{eq:linModel} is that the input is a 1D vector. Meaning, the 2D image is flattened into a 1D vector. For tasks like image segmentation, spatial structure of the images can be informative in making improved segmentation decisions. Loss of spatial pixel correlation can be detrimental to downstream tasks; more so when dealing with complex structures encountered in medical images.
\end{enumerate}
%  \\ \\
%
We overcome these two constraints by approximating the linear model in Equation~\eqref{eq:linModel} using MPS tensor network and by operating on strides of smaller non-overlapping image patches. 
\\
\\
{\bf Matrix Product State} 
%As discussed above, the weight tensor, $\Theta$, in Equation~\eqref{eq:linModel} has $d^N$ tunable parameters.  
Computing the inner product in Eq.~\eqref{eq:linModel} becomes infeasible with increasing $N$~\cite{stoudenmire2016supervised}. It also turns out that only a small number of degrees of freedom in these exponentially high dimensional Hilbert spaces are relevant~\cite{poulin2011quantum,orus2014practical}. These relevant degrees of freedom can be efficiently accessed using tensor networks such as the MPS~\cite{perez2006matrix,oseledets2011tensor}. In image analysis,  accessing this smaller sub-space of the high dimensional Hilbert space corresponds to accessing interactions between pixels that are local either in spatial- or in some feature-space sense that is relevant for the task.
%The $d^N$ lift in quantum physics is sometimes seen as a convenient illusion~\cite{poulin2011quantum,orus2014practical} as the most interesting behavior of the systems can be captured with fewer degrees of freedom in this high dimensional space.

%The inner product in Equation~\eqref{eq:linModel} can be approximated using the MPS tensor network~\cite{perez2006matrix,oseledets2011tensor}. 

\begin{wrapfigure}{r}{0.45\textwidth}
    % \vspace{-0.25cm}
    \centering
    \includegraphics[width=0.45\textwidth]{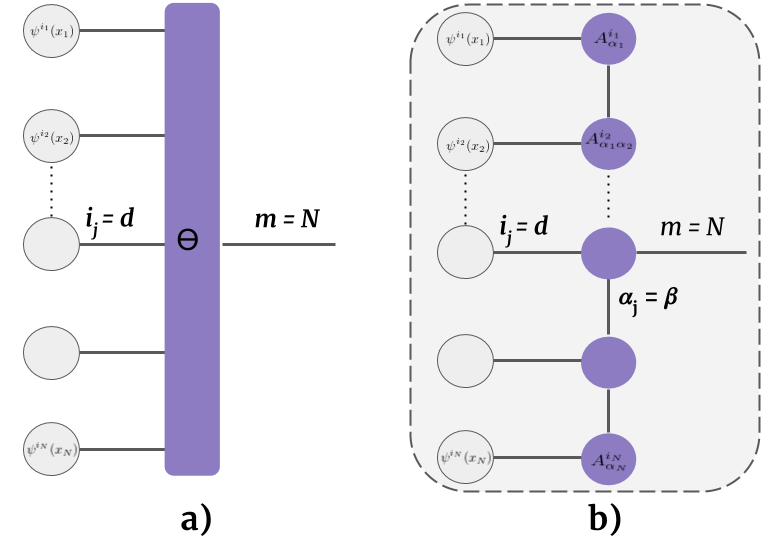}
    \caption{  {\bf a)} Linear decision rule in Equation~\ref{eq:linModel} depicted in tensor notation. Note that $\Theta$ has N+1 edges as it is an order-(N+1) tensor. The d-dimensional local feature maps are the gray nodes marked $\psi^{i_j}(x_j)$. {\bf b)} Matrix product state (MPS) approximation of $\Theta$ in Equation~\ref{eq:mps} into a tensor train comprising up to order-3 tensors, $A^{i_j}_{\alpha_j\alpha_{j+1}}$.}
    \label{fig:mps}
    \vspace{-0.5cm}
\end{wrapfigure}

MPS is a tensor factorisation method that can approximate any order-N tensor with a chain of order-3 tensors. This is visualized using tensor notation in Figure~\ref{fig:mps}-b for approximating $\Theta^{m}_{i_1\dots i_N}$ using $A^{i_j}_{\alpha_j \alpha_{j+1}} \forall\; j=1\dots N$ which are of order-3 (except on the borders where they are order-2). The dimension of subscript indices of  ${\alpha_j}$ which are contracted can be varied to yield better approximations. These variable dimensions of the intermediate tensors in MPS are known as bond dimension $\beta$. MPS approximation of $\Theta^{m}_{i_1\dots i_N}$ depicted in Figure~\ref{fig:mps}-b is given by
\begin{equation}
    \Theta^{m}_{i_1\dots i_N} = \sum_{\alpha_1, \alpha_2,\dots \alpha_N} A^{i_1}_{\alpha_1} A^{i_2}_{\alpha_1 \alpha_2} A^{i_3}_{\alpha_2 \alpha_3} \dots A^{m,i_j}_{\alpha_j \alpha_{j+1}} \dots A^{i_N}_{\alpha_N}.
    \label{eq:mps}
\end{equation}
The components of these intermediate lower-order tensors $A^{i_j}_{\alpha_j \alpha_{j+1}} \forall\; j=1\dots N$ form the tunable parameters of the MPS tensor network. 
This MPS factorisation in Equation~\eqref{eq:mps} reduces the number of parameters to represent $\Theta$ from $N\cdot d^N$ to  $\{N \cdot d \cdot N \cdot \beta^2\}$ with $\beta$ controlling the quality of these approximations\footnote{Tensor indices are dropped for brevity in the remainder of the manuscript.}. Note that when $\beta=d^{N/2}$ the MPS approximation is exact~\cite{orus2014practical,stoudenmire2016supervised}. 

Various implementations of MPS perform the sequence of tensor contractions in different ways for efficiency. We use the TorchMPS implementation in this work, which first contracts the horizontal edges and performs tensor contractions along the vertical edges~\cite{torchmps,raghav2020tensor}.
\\
\\
%where  $A^{i_j}$ are the lower-order tensors. The subscript indices $\alpha_j$ are the virtual indices that are contracted. Dimension of the virtual indices is $\beta$ which is the \emph{bond dimension}. 
%shows the tensor contraction based approximation to the dot product using tensor notations~\cite{penrose1971applications}. Notice that the chain of tensors approximating $\Theta$ are at most of order-3. The edges connecting $A_$
%The MPS  approximation in tensor notation is indicated as the {\em contract} step in Figure~\ref{fig:lotenet}. Tensor indices are dropped in the remainder of the manuscript for ease of notation.
% Additionally, we propose to learn $f_\Theta (\cdot)$ that operate on smaller regions of the input image. This has two advantages: 1) Reduction in number of parameters due to weight sharing 2) Operating on smaller regions 
{\bf MPS on non-overlapping patches} The issue of loss in spatial pixel correlation is not alleviated with MPS as it operates on flattened input images. MPS with higher bond dimensions could possibly allow interactions between all pixels but, due to the quadratic increase in number of parameters with the bond dimension $\beta$, working with higher bond dimensions can be prohibitive.

To address this issue, we apply MPS on small non-overlapping image regions. These smaller patches can be flattened without severe degradation of  spatial correlation. Similar strategy of using MPS on small image regions has been used for image classification using tensor networks in~\cite{raghav2020tensor}. This is also in the same spirit of using convolutional filter kernels in CNNs when the kernel width is set to be equal to the stride length. This formulation of using MPS on regions of size $K\times K$ with a stride equal to $K$ in both dimensions results in the strided tensor network formulation, given as
\begin{align}
f(\xbf;\Theta_K) &= \{ \Theta_K \cdot \Phi( \xbf_{(i,j)})\} \quad \forall\; i = 1,\dots,H/K, \; j = 1,\dots,W/K
% \nonumber \\ 
% & \triangleq \Theta_K \boxdot  \Phi(\xbf)
    \label{eq:model}    
\end{align}
where $(i,j)$ are used to index the patch from row $i$ and column $j$ of the image grid with patches of size $K\times K$. The weight matrix in Equation~\eqref{eq:model}, $\Theta_K$ is subscripted with $K$ to indicate that MPS operations are performed on $K\times K$ patches.

In summary, with the proposed strided tensor network formulation, linear segmentation decisions in Equation~\eqref{eq:linModel} are approximated at the patch level using MPS. The resulting patch level predictions are tiled back to obtain the $H\times W$ segmentation mask. 

% \begin{figure}[t]
%     % \centering
%     \begin{minipage}{0.4\textwidth}
% % \begin{figure}
%     \centering
%     \includegraphics[width=0.79\textwidth]{figures/tensorNotation.png}
% % \end{figure}
% \end{minipage}  
% \begin{minipage}{0.59\textwidth}
% % \begin{figure}
%     \centering
%     \includegraphics[width=0.99\textwidth]{figures/mps.png}
% % \end{figure}
% \end{minipage}  
% \caption{  c) Linear decision rule in Equation~\ref{eq:linModel} depicted in tensor notation. Note that $\Theta$ has N+1 edges as it is an order-(N+1) tensor. The d-dimensional local feature maps are the gray nodes marked $\psi^{i_j}(x_j)$. d) Matrix product state (MPS) approximation of $\Theta$ in Equation~\ref{eq:mps} into a tensor train comprising order-3 or order-3 tensors, $A^{i_j}_{\alpha_j\alpha_{j+1}}$.}
%     \label{fig:tensor}
%     \label{fig:my_label}
% \end{figure}
\vspace{-0.25cm}
\subsection{Optimisation} 

The weight tensor, $\Theta_K$, in Equation~\eqref{eq:model} and in turn the lower order tensors in Equation~\eqref{eq:mps} which are the model parameters can be learned in a supervised setting. For a given labelled training set with $T$ data points, $\mathcal{D}:\{(\xbf_1,\ybf_1), \dots (\xbf_T,\ybf_T)\}$, the training loss to be minimised is
\begin{equation}
    \mathcal{L}_{tr} = \frac{1}{T} \sum_{t=1}^T L(f(\xbf_i),\ybf_i),
\end{equation}
where $\ybf_i$ are the binary ground truth masks and $L(\cdot)$ can be a suitable loss function suitable for segmentation tasks. In this work, as both datasets were largely balanced (between foreground and background classes) we use binary cross entropy loss. 

% \begin{figure}[t]
%     \centering
%     \includegraphics[width=0.6\textwidth]{figures/mps.png}
%     \caption{MPS approximation}
%     \label{fig:my_label}
% \end{figure}
\vspace{-0.25cm}
\section{Data \& Experiments}
\vspace{-0.25cm}
\label{sec:data_exp}
Segmentation performance of the proposed strided tensor network is evaluated on two datasets and compared with relevant baseline methods. Description of the data and the experiments are presented in this section.
\vspace{-0.25cm}
\subsection{Data}
\label{sec:data}
\vspace{-0.25cm}
\begin{figure}[t]
    \centering
    \includegraphics[width=0.95\textwidth]{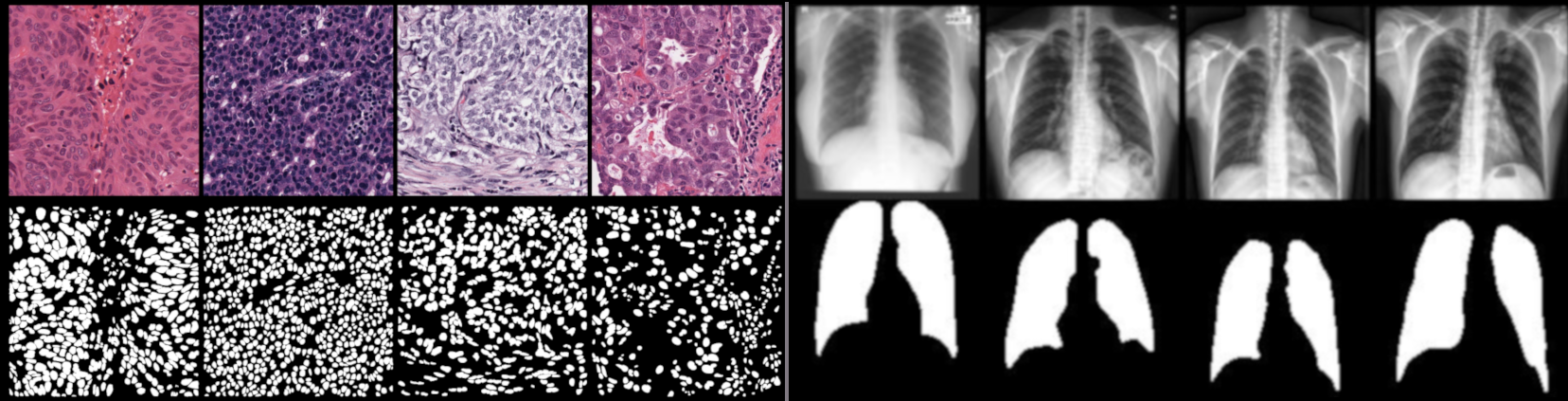}
    \vspace{-0.25cm}
    \caption{(First four columns) Sample images from the MO-Nuseg dataset comprising histopathology slides  from multiple organs (top row) and the corresponding binary masks (bottom row). (Last four columns) Sample chest X-ray images from the Lung-CXR dataset with corresponding binary masks. }
    \label{fig:monuseg}
    \vspace{-0.5cm}
\end{figure}

% \subsubsection
{\bf MO-NuSeg Dataset}
The first dataset we use in our experiments is the multi-organ nuclei segmentation (MO-NuSeg) challenge dataset\footnote{\url{https://monuseg.grand-challenge.org/}}~\cite{kumar2017dataset}. This dataset consists 44 Hematoxylin and eosin (H\&E) stained tissue images, of size 1000$\times$1000, with about 29,000 manually annotated nuclear boundaries. The dataset has tissues from seven different organs and is a challenging one due to the variations across different organs. The challenge organizers provide a split of 30 training/validation images and 14 testing images which we also follow to allow for comparisons with other reported methods. Four training tissue images and the corresponding binary nuclei masks are shown in the first four columns of Figure~\ref{fig:monuseg}.
\vspace{0.2cm} \\
% \subsubsection
{\bf Lung-CXR Dataset}
\label{sec:lungCXR}
We also use the lung chest X-ray dataset collected from the Shenzhen and Montgomery hospitals with posterio-anterior views for tuberculosis diagnosis~\cite{jaeger2014two}. The CXR images used in this work are of size 128$\times$128 with corresponding binary lung masks for a total of 704 cases which is split into training ($352$), validation ($176$) and test ($176$) sets. Four sample CXRs and the corresponding binary lung masks are shown in the last four columns of Figure~\ref{fig:monuseg}.
\vspace{-0.25cm}
\subsection{Experiments}
\label{sec:exp}
\vspace{-0.25cm}
% \subsubsection
{\bf Experimental set-up} The proposed strided tensor network model was compared with a convolutional neural network (U-net~\cite{ronneberger2015u}), a modified tensor network (MPS TeNet) that uses one large MPS operation similar to the binary classification model in~\cite{efthymiou2019tensornetwork}, and a multi-layered perceptron (MLP). 
% The initial feature map 
% U-net model has an initial feature map of $8$ which is doubled in each of the five resolutions in the encoding path. The baseline model MPS TeNet uses a bond dimension, $\beta=10$~\cite{efthymiou2019tensornetwork}. MLP comprises of 6 levels with ReLU non-linear activations. 
Batch size of $1$ and $32$ were used for the MO-NuSeg and Lung-CXR datasets, respectively, which was the maximum batch size usable with the baseline U-net; all other models were trained with the same batch size for a fairer comparison. The models were trained with the Adam optimizer and an initial learning rate of $5\times 10^{-4}$, except for the MPS TeNet which required smaller learning rate of $1\times 10^{-5}$ for convergence; they were trained until there was no improvement in validation accuracy for 10 consecutive epochs. The model based on the best validation performance was used to predict on the test set. All models were implemented in PyTorch and trained on a single GTX 1080 graphics processing unit (GPU) with $8$GB memory. The development and training of all models in this work was estimated to produce $61.9$ kg of CO2eq, equivalent to $514.6$ km travelled by car as measured by Carbontracker\footnote{\url{https://github.com/lfwa/carbontracker/}}~\cite{anthony2020carbontracker}.
\\
{\bf Metrics} Performance of the different methods for both datasets are compared using Dice score based on binary predictions, $\hat{y}_i\in\{0,1\}$ obtained by thresholding soft segmentations at $0.5$ which we recognise is an arbitrary threshold. A more balanced comparison is provided using the area under the precision-recall curve (PRAUC or equivalently the average precision) using the soft segmentation predictions, ${s}_i\in[0,1]$. 
\\
{\bf Model hyperparameters } The initial number of filters for the U-net model was tuned from $[8,16,32,64]$ and a reasonable choice based on validation performance and training time was found to be $8$ for MO-NuSeg dataset and $16$ for Lung-CXR dataset. The MLP consists of 6 levels, 64 hidden units per layer and ReLU activation functions and was designed to match the strided tensor network in number of parameters. The strided tensor network has two critical hyperparameters: the bond dimension ($\beta$) and the stride length ($K$), which were tuned using the validation set performance. The bond dimension controls the quality of the MPS approximations and was tuned from the range $\beta=[2,4,8,12,16,20,24]$. The stride length controls the field of view of the MPS and was tuned from the range $K=[2,4,8,16,32,64,128]$. 
%We fixed $\beta=24$ as higher bond dimensions yield better approximations of $\Theta$, and only tuned the stride parameter; using the best stride parameter influence of other bond dimensions were evaluated. As the range of parameters are not massive (49 combinations) one could also perform a simple grid search. 
For MO-NuSeg dataset, the best validation performance was stable for any $\beta\geq4$, so we used the smallest with $\beta=4$ and the best performing stride parameter was $K=8$. For the Lung-CXR dataset similar performance was observed with $\beta\geq 20$, so we set $\beta=20$ and obtained $K=32$. The local feature dimension ($d$) was set to $4$ (see Section~\ref{sec:disc} for additional discussion on local feature maps). 

\begin{table}[t]
\small
\centering
    \caption{ Test set performance comparison for segmenting nuclei from the stained tissue images (MO-NuSeg) and segmenting lungs from chest CT (Lung-CXR). For all models, we report the number of parameters $|\Theta|$, computation time per training epoch, area under the curve of the precision-recall curve (PRAUC) and average Dice accuracy (with standard deviation over the test set). The representation (Repr.) used by each of the methods at input is also mentioned.}
    % Best performing PR-AUC and Dice across all models are shown in boldface.}
    \label{tab:res}
     \centering
     \vspace{-0.15cm}
  \begin{tabular}{clcrrrccc}
    \toprule
    {\bf Dataset} & {\bf Models} & {\bf Repr.} &&{\bf $|\mathbf{\Theta|}$}   && {\bf t(s)} & {\bf PRAUC} & {\bf Dice} \\
    \midrule
    \multirow{4}{*}{\bf MO-NuSeg}& Strided TeNet (ours) &1D && $5.1K$&&$21.2$&$0.78$&$0.70\pm 0.10$     \\
    & U-net~\cite{ronneberger2015u} &2D&&$500K$&& $24.5$&$0.81$&    $0.70\pm0.08$\\
    & MPS TeNet~\cite{efthymiou2019tensornetwork} &1D&&$58.9M$ && $240.1$ &$0.55$&$0.52\pm0.09$\\
    & CNN  &2D&& -- && $510$ & -- &$0.69\pm 0.10 $\\
    \midrule
    \midrule
    \multirow{4}{*}{\bf Lung-CXR}&Strided TeNet (ours) & 1D&&$2.0M$&&$6.1$&$0.97$&$0.93\pm0.06$     \\
    &U-net~\cite{ronneberger2015u} &2D&&$4.2M$&& $4.5$&$0.98$&    $0.95\pm0.02$\\
    &MPS TeNet~\cite{efthymiou2019tensornetwork}&1D && $8.2M$&&$35.7$ &$0.67$&$0.57\pm 0.09$\\
    &MLP &1D&& $2.1M$ && $4.1$&$0.95$&$0.89\pm0.05$\\
    \bottomrule
  \end{tabular}
  \vspace{-0.25cm}
\end{table}
% \footnotetext[7]{These numbers are reported from~\cite{kumar2017dataset} for their CNN2 model used for binary segmentation. Run time could be shorter with more recent hardware.}
% \addtocounter{footnote}{7}
\vspace{-0.25cm}
\subsection{Results}
\begin{figure}[t]
    % \vspace{-0.1cm}
    \centering
    \includegraphics[width=0.65\textwidth]{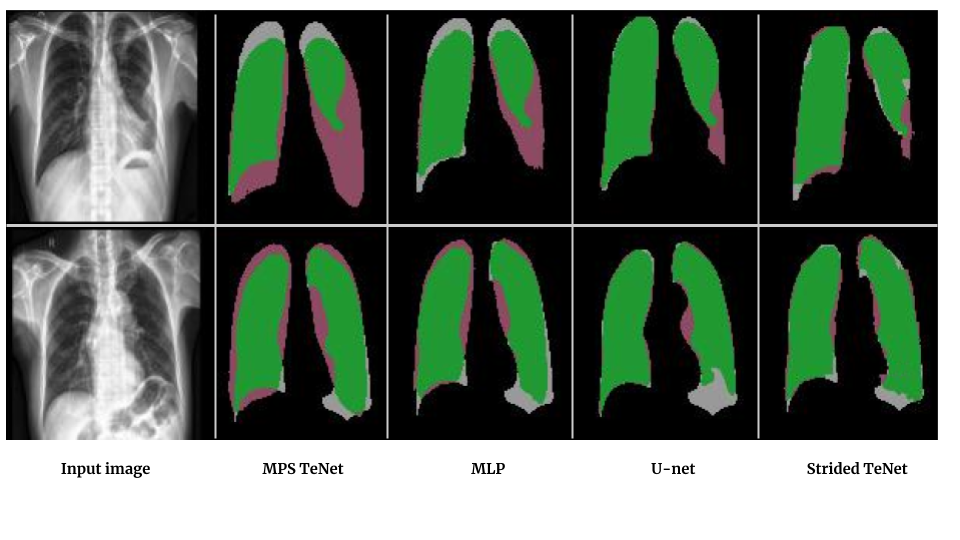}
    \vspace{-0.5cm}
    \caption{Two test set CXRs from Lung-CXR dataset along with the predicted segmentations from the different models. All images are upsampled for better visualisation. (Prediction Legend -- Green: True Positive, Grey: False Negative, Pink: False Positive) }
    \label{fig:cxr_res}
    \vspace{-0.5cm}
\end{figure}

{\bf MO-NuSeg} Performance of the strided tensor network compared to the baseline methods on the MO-NuSeg dataset are presented in Table~\ref{tab:res} where we report the PRAUC, Dice accuracy, number of parameters and the average training time per epoch for all the methods. Based on both PRAUC and Dice accuracy, we see that the proposed strided tensor network (PRAUC=$0.78$, Dice=$0.70$) and the U-net (PRAUC=$0.81$, Dice=$0.70$) obtain similar performance. There was no significant difference between the two methods based on paired sample t-tests. The Dice accuracy reported in the paper that introduced the dataset\footnote{These numbers are reported from~\cite{kumar2017dataset} for their CNN2 model used for binary segmentation. Run time in Table~\ref{tab:res} for CNN2 model could be lower with more recent hardware.} in~\cite{kumar2017dataset} ($0.69$) is also in the same range as the reported numbers for the strided tensor network. A clear performance difference is seen in comparison to the MPS tensor network (PRAUC=$0.55$, Dice=$0.52$); the low metrics for this method is expected as it is primarily a classification model~\cite{efthymiou2019tensornetwork} modified to perform segmentation to compare with the most relevant tensor network. 
% Two test samples and the predicted segmentations are shown in Figure~\ref{fig:nuseg_res}. 
\vspace{0.02cm}
\\
{\bf Lung-CXR dataset } Segmentation accuracy of the strided tensor network is compared to U-net, MPS tensor network and MLP on the Lung-CXR dataset in Table~\ref{tab:res}. U-net ($0.98$) and the proposed strided tensor network ($0.97$) attain very high PRAUC, and the Dice accuracy for strided tensor network is $0.93$ and for U-net it is $0.95$; there was no significant difference based on paired sample t-test. Two test set predictions where the strided tensor network had high false positive (top row) and high false negative (bottom row), along with the predictions from other methods and the input CXR are shown in Figure~\ref{fig:cxr_res}. 
\vspace{-0.25cm}

% \begin{table}[t]
% % \tiny
% \centering
%     \caption{Lung-CXR test set performance comparison for segmenting lungs from chest CT. For all models, we report the number of parameters $|\Theta|$, average training time per epoch, area under the precision-recall curve (PRAUC) and Dice accuracy. The representation (Repr.) used by each of the methods at input is also mentioned.}
%     % Best performing PR-AUC and Dice across all models are shown in boldface.}
%     \label{tab:res}
%      \centering
%     \begin{tabular}{lcrrrrccc}
%     % \vspace{1cm}
%     \toprule
%     Models & Repr. &&$|\Theta|(M)$  && t(s) & PRAUC & Dice \\
%     \midrule
%     Strided TeNet (ours) & 1D&&$2.0$&&$6.1$&$0.97$&$0.93\pm$     \\
%     U-net~\cite{ronneberger2015u} &2D&&$4.2$&& $4.5$&$0.98$&    $0.95\pm$\\
%     MPS TeNet~\cite{efthymiou2019tensornetwork}&1D && $8.2$&&$35.7$ &$0.67$&$0.57\pm$\\
%      MLP &1D&& $2.1$ && $4.1$&$0.95$&$0.89\pm$\\
%     \bottomrule
%     \end{tabular}
% \end{table}
% \begin{figure}
%     \centering
%     \includegraphics[width=0.6\textwidth]{figures/nuseg_res.jpg}
%     \caption{Two test set slides from MO-NuSeg dataset along with the predicted segmentations from the different models. (Prediction Legend -- Green: True Positive, Grey: False Negative, Pink: False Positive) }
%     \label{fig:nuseg_res}
% \end{figure}
%
% MONSeg: $\beta=20,d=8,B=1$, dice loss, U-net
% MONSeg: $\beta=5,d=4,B=1$, BCE, U-net
% Lung CXR: $\beta=20,d=8,B=32$, BCE, U-net
% Lung CXR: $\beta=12,d=4,B=32$, BCE, U-net

% Show the worst case predictions from both models; discuss the mistakes. 

\vspace{-0.1cm}
\section{Discussion \& Conclusions}
\label{sec:disc}
\vspace{-0.1cm}
% {\bf Discussion of results}
Results from Table~\ref{tab:res} show that the proposed strided tensor network compares favourably to other baseline models across both datasets. In particular, to the U-net where there is no significant difference in Dice accuracy and PRAUC. The computation cost per training epoch of strided tensor network is also reported in Table~\ref{tab:res}. The training time per epoch for both datasets for the proposed strided tensor network is in the same range of those for U-net. In multiple runs of experiments we noticed that U-net converged faster ($\approx 50$ epochs) whereas the tensor network model converged around $80$ epochs. Overall, the training time for tensor network models on both datasets was under one hour. A point also to be noted is that operations for CNNs are highly optimised in frameworks such as PyTorch. More efficient implementations of tensor network operations are being addressed in recent works~\cite{fishman2020itensor,novikov2020tensor} and the computation cost is expected to reduce further.

An additional observation in the experiments on MO-NuSeg data, reported in Table~\ref{tab:res} is the number of parameters used by the strided tensor network ($5.1K$) which is about two orders of magnitude smaller than that of the U-net ($500K$), without a substantial difference in segmentation accuracy. As a consequence, the maximum GPU memory utilised by the proposed tensor network was $0.8$GB and it was $6.5$GB for U-net. This difference in GPU memory utilisation can have ramifications as the strided tensor network can handle larger batch sizes (resulting in more stable, faster training).
%or fit models with higher parameters without incurring additional hardware costs when compared to CNNs. 
This could be most useful when dealing with large medical images which can be processed without patching them for processing.
%can also be a useful feature when working with large images encountered in medical imaging tasks which can be without having to patch them. 
In general, tensor network models require lesser GPU memory as they do not have intermediate feature maps and do not store the corresponding computation graphs in memory, similar to MLPs~\cite{raghav2020tensor}.
%We evaluated this performance by trying to train strided tensor network with the largest batch size possible:  for the MO-NuSeg where images were of size $1000\times1000$, it was $B=12$ (whereas $B=1$ for U-net) without any degradation in performance.

\begin{figure}[t]
    \centering
    \vspace{-0.25cm}
    \includegraphics[width=0.65\textwidth]{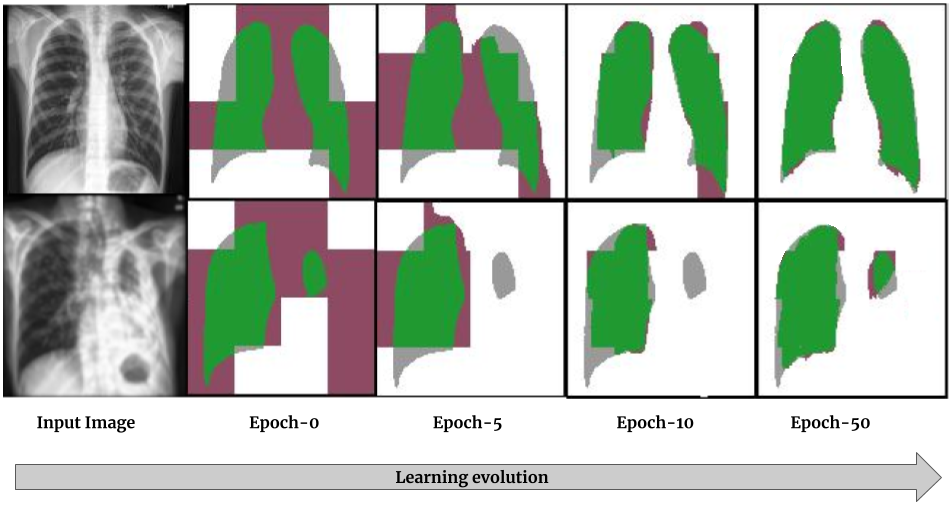}
    \vspace{-0.35cm}
    \caption{Progression of learning for the strided tensor network, for the task of segmenting lung regions. Two {\em validation} set input chest CT images (column 1) and the corresponding predictions at different training epochs are visualized overlaid with the ground truth segmentation. Images are of size 128$\times$128 and the stride is over 32$\times$32 regions. All images are upsampled for better visualisation. (Prediction Legend -- Green: True Positive, Grey: False Negative, Pink: False Positive) }
    \label{fig:evolution}
    \vspace{-0.5cm}
\end{figure}

The predicted lung masks in Figure~\ref{fig:cxr_res} show some interesting underlying behaviours of the different methods. MPS TeNet, MLP and the strided tensor network all operate on 1D representations of the data (flattened 2D images). The influence of loss of spatial correlation due to flattening is clearly noticeable with the predicted segmentations from MPS TeNet (column 2) and MLP (column 3), where both models predict lung masks which resemble highly regularised lung representations learned from the training data. The predictions from strided tensor network (column 5) are able to capture the true shape with more fidelity and are closer to the U-net (column 4).  This behaviour could be attributed to the input representations used by each of these models. U-net operates on 2D images and does not suffer from this loss of spatial correlation between pixels. The proposed strided tensor network also operates on flattened 1D vectors but only on smaller regions due to the patch-based striding. Loss of spatial correlation in patches is lower when compared to flattening full images. 
\vspace{0.2cm}\\ 
{\bf Influence of local feature maps}
Local feature map in Equation~\eqref{eq:local} used to increase the local dimension ($d$) are also commonly used in many kernel based methods and these connections have been explored in earlier works~\cite{stoudenmire2016supervised}. We also point out that the local maps are similar to {\em warping} in Gaussian processes which are used to obtain non-stationary covariance functions from stationary ones~\cite{mackay1998introduction}. While in this work we used simple sinusoidal transformations, it has been speculated that more expressive features could improve the tensor network performance~\cite{stoudenmire2016supervised}. To test this, we explored the use of jet-based features~\cite{larsen2012jet} which can summarise pixel neighbourhoods using first and second order derivatives at multiple scales, and have previously shown promising results. However, our experiments on the Lung-CXR dataset showed no substantial difference in the segmentation performance compared to when simply using the sinusoidal feature map in Equation~\eqref{eq:local}. We plan on exploring a more comprehensive analysis of local feature maps on other datasets in possible future work.
% Further investigation on trying out other features such as wavelet-based ones~\cite{reyes2020multi} could be interesting. 
% \\ \\
%
% {\bf Trade off between stride and bond dimension}
% 
%This was also shown to be beneficial for image classification tasks
\vspace{0.25cm}
\\ 
{\bf Learning a single filter?} 
CNNs like the U-net thrive by learning banks of filters at multiple scales, which learn distinctive features and textures, making them effective in computer vision tasks. The proposed strided tensor network operates on small image patches at the input resolution. Using the CNN framework, this is analogous to learning a single filter that can be applied to the entire image. We illustrate this behaviour in Figure~\ref{fig:evolution}, where the strided tensor network operates on $32\times32$ image regions. We initially see block effects due to the large stride and within a few epochs the model is able to adapt to the local pixel information. This we find to be quite an interesting behaviour. A framework that utilises multiple such tensor network based {\em filters} could make these classes of models more versatile and powerful.
% \vspace{-0.25cm}
%
\vspace{0.25cm}
\\ %\\
% \section{Conclusions}
% \vspace{-0.25cm}
{\bf Conclusions}
In this work, we presented a novel tensor network based image segmentation method. We proposed the strided tensor network which uses MPS to approximate hyper-planes in high dimensional spaces, to predict pixel classification into foreground and background classes. In order to alleviate the loss of spatial correlation, and to reduce the exponential increase in number of parameters, the strided tensor network operates on small image patches. We have demonstrated promising segmentation performance on two biomedical imaging tasks. The experiments revealed interesting insights into the possibility of applying linear models based on tensor networks for image segmentation tasks. This is a different paradigm compared to the CNNs, and could have the potential to introduce a different class of supervised learning methods to perform image segmentation. 

\vspace{-0.2cm}
\tiny
\bibliographystyle{splncs04}
\vspace{-0.25cm}
\bibliography{M335.bib}
% \vspace{-0.5cm}

\end{document}